\documentclass[10pt,twocolumn,letterpaper]{article}

\usepackage{iccv}
\usepackage{times}
\usepackage{epsfig}
\usepackage{graphicx}
\usepackage{amsmath}
\usepackage{amssymb}
\usepackage{graphics}
\usepackage{multirow}
\usepackage{booktabs}
\usepackage{enumitem}
\usepackage{algorithm}
\usepackage{algpseudocode}

\newcommand{\tabincell}[2]{\begin{tabular}{@{}#1@{}}#2\end{tabular}} 
\usepackage{amsfonts,bm}
\usepackage{verbatim}

\usepackage{etoolbox}
\makeatletter
\patchcmd{\maketitle}
 {\def\@makefnmark}
 {\def\@makefnmark{}\def\useless@macro}
 {}{}
\makeatother

\def\vf{{\bm{f}}}
\def\vg{{\bm{g}}}
\def\vh{{\bm{h}}}

\def\vl{{\bm{l}}}

\def\mF{{\bm{F}}}

\def\gI{{\mathcal{I}}}

\def\gO{{\mathcal{O}}}

\def\sD{{\mathbb{D}}}

\usepackage[breaklinks=true,letterpaper=true,colorlinks,bookmarks=false]{hyperref}

\iccvfinalcopy 

\def\iccvPaperID{491}

\ificcvfinal\fi

\begin{document}

\title{Auto-ReID: Searching for a Part-Aware ConvNet for Person Re-Identification}

\author{Ruijie Quan$^{1,2}$, Xuanyi Dong$^{1,2}$, Yu Wu$^{1,2}$, Linchao Zhu$^{2}$, Yi Yang$^{2}$\\
$^{1}$Baidu Research\thanks{{Part of this work was done when Ruijie Quan, Xuanyi Dong, and Yu Wu interned at Baidu Research. Yi Yang is the corresponding author.}}~~~~~~$^{2}$ReLER, University of Technology Sydney\\
{\tt\small \{ruijie.quan,xuanyi.dong,yu.wu-3\}@student.uts.edu.au} \\
{\tt\small \{linchao.zhu,yi.yang\}@uts.edu.au}
}

\maketitle
\ificcvfinal\thispagestyle{empty}\fi
\graphicspath{image}

\begin{abstract}

Prevailing deep convolutional neural networks (CNNs) for person re-IDentification (reID) are usually built upon ResNet or VGG backbones, which were originally designed for classification. Because
reID is different from classification, the architecture should be modified accordingly.
We propose to automatically search for a CNN architecture that is specifically suitable for the reID task.
There are three aspects to be tackled. 
First, body structural information plays an important role in reID but it is not encoded in backbones.
Second, Neural Architecture Search (NAS) automates the process of architecture design without human effort, but no existing NAS methods incorporate the structure information of input images.
Third, reID is essentially a retrieval task but current NAS algorithms are merely designed for classification.
To solve these problems, we propose a retrieval-based search algorithm over a specifically designed reID search space, named Auto-ReID.
Our Auto-ReID enables the automated approach to find an efficient and effective CNN architecture for reID. Extensive experiments demonstrate that the searched architecture achieves state-of-the-art performance while reducing 50\% parameters and 53\% FLOPs compared to others.

\end{abstract}

\section{Introduction}

Person re-IDentification (reID) aims to retrieve the images of a person recorded by different surveillance cameras~\cite{li2014deepreid,zheng2015scalable}.
With the success of deep convolutional neural networks (CNNs) in recent years, researchers in this area have mainly focused on improving the representation capability of the features extracted from CNN models~\cite{Su_2017_PDC,zheng2015scalable,Sun_2018_PCB}.
Hundreds of different CNN models have been designed for reID, and the rank-1 accuracy has been improved from 44.4\%~\cite{zheng2015scalable} to 93.8\%~\cite{Sun_2018_PCB} on the Market-1501 benchmark~\cite{zheng2015scalable}.

\begin{figure}[!t]
\label{fig:our method}
\centering
\includegraphics[width=\linewidth]{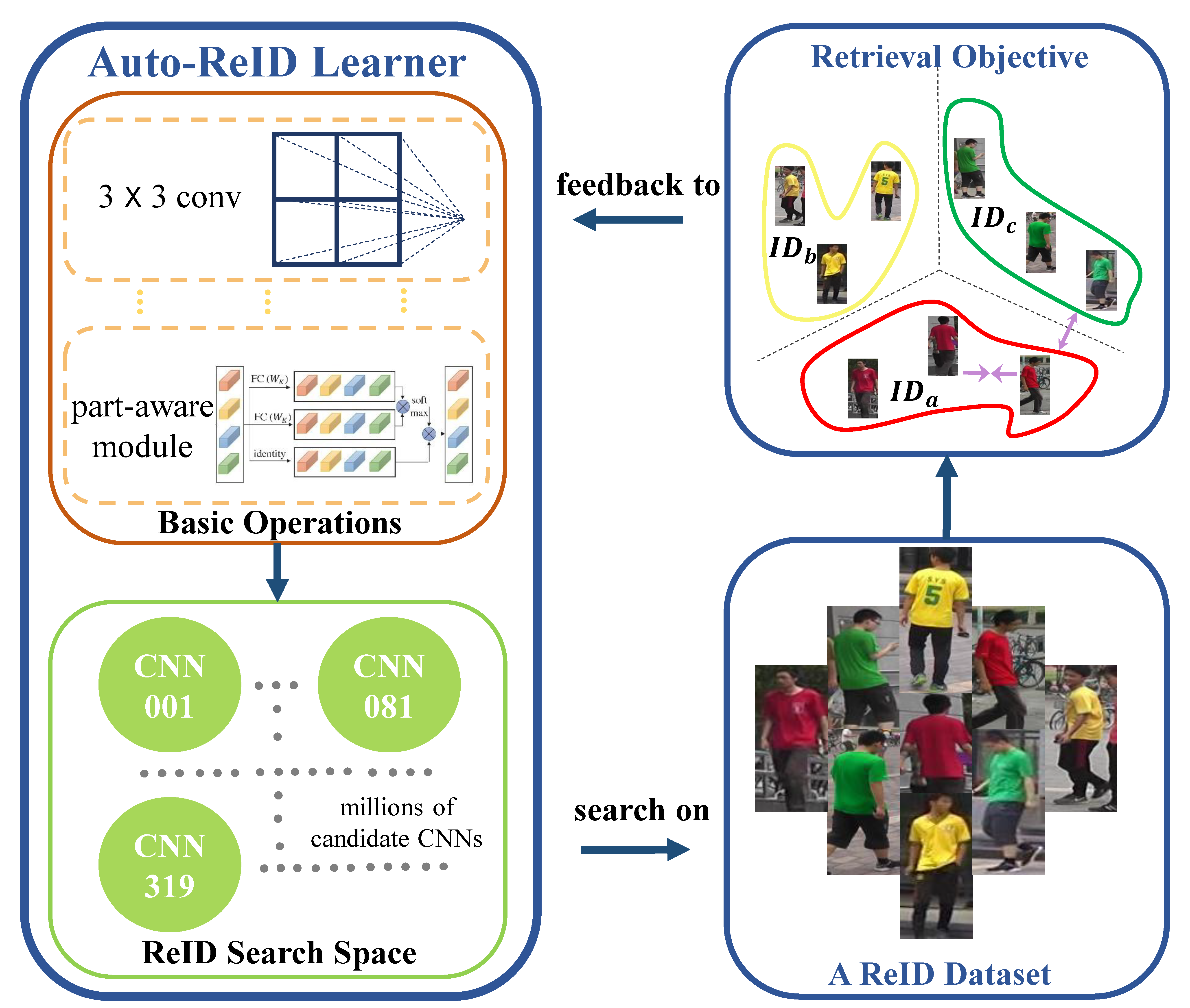}
\caption{
Our Auto-ReID learns to search for a suitable architecture on a specific reID dataset, and it is supervised by the retrieval objective during searching.
Auto-ReID finds architecture from a reID search space, which consists of a large number of candidate architectures. These candidates are generated by combining basic operations, such as a 3-by-3 convolutional layer, a 3-by-3 max pooling operation, and the proposed part-aware module.
}
\end{figure}

Most recent reID models are based on deep CNNs. They are usually built upon convolutional neural network backbones for image classification~\cite{xiao2016learning,li2014deepreid,Su_2017_PDC,suh2018part,Sun_2018_PCB}, such as VGG~\cite{simonyan2015very}, Inception~\cite{Szegedy_2015_googlenet}, and ResNet~\cite{he2016resnet}.
These backbones can be readily used for retrieval as the inputs of both tasks are images. However, there are still a few differences between the reID task and the classification task. 
For example, in image classification, the appearance of two objects could be different, e.g., a cat looks different from a tree. In contrast, all input examples of the reID task are person images with different attributes, e.g., apparel or hair styles. 
A CNN focusing on recognizing over 1,000 objects~\cite{deng2009imagenet} should be modified when it is applied to the reID task.

A straightforward method is to manually design a reID oriented CNN architecture which is specifically suitable for the reID problem. 
However, manually designing an exquisite architecture for the reID task may take months~\cite{ZophL2017NASNet,Hieu2018ENAS,liu2019darts} even for human experts. This is inefficient and labor intensive.
In this paper, we propose an automated approach to search for an optimal CNN architecture that is explicitly suited to the reID task.
Our premise is that the CNN backbones designed for classification may have redundant and missing components for retrieval (the reID task), e.g., (1) less pooling layers benefit to reID accuracy and (2) no component for classification explicitly captures body structure information.
There remain three challenges to automate Neural Architecture Search (NAS) for reID.
First, no existing NAS approaches search for a CNN architecture that preserves body structural information.
The body structure information plays an important role in reID, which is a major difference between reID and classification~\cite{suh2018part,li2014deepreid}.
Second, reID methods usually encode structural information in a backbone-dependent way. They require extensive manual tuning of the hyper-parameters when a different backbone network is adopted~\cite{Sun_2018_PCB,saquib2018pose}.
Third, reID is essentially a retrieval task, but most NAS algorithms are designed for classification. Since retrieval and classification have different objectives, existing NAS algorithms are not directly applicable to the reID problem~\cite{ZophL2017NASNet,Hieu2018ENAS,Liu_2018_PNAS,liu2019darts}.

In this paper, we propose an approach called Auto-ReID to solve these three challenges. 
The key contribution of Auto-ReID lies in the design of a new reID search space. This design enables us to construct more optimal architectures which make best use of human body structure information.
Specifically, we design a part-aware module to enhance the body structure information of a given input feature tensor.
Unlike existing part-based reID models, the proposed part-aware module is flexible and able to handle features with various input shapes. We use this module as a basic operation for constructing a number of reID candidate architectures. 
In addition to a typical softmax loss, the proposed Auto-ReID equips the differentiable NAS method ~\cite{liu2019darts} with a retrieval loss, making the search results particularly suitable for the reID task. 
The combination of the proposed reID search space and reID search algorithm enables us to find an efficient and effective architecture for reID in an automated way (Fig.~\ref{fig:our method}). In summary, our contributions are as follows:
\begin{itemize}
    \setlength\itemsep{0em}
    \item This is the first approach that searches neural architectures for the reID task, eliminating human experts' effort in the manual design of CNN models for reID.
    \item We propose a novel reID search space in which body structure is formulated as a trainable and operational CNN component. The proposed reID search space combines (1) modules that explicitly capture pedestrian body part information and (2) typical modules that have been used in the standard NAS search space. 
    \item We integrate a retrieval loss into the differentiable NAS algorithm so as to better fit the reID task. 
    We adopt the modified searching strategy and batch data sampling method in accordance with the new retrieval objective.
    \item Extensive experiments show that the searched CNN achieves competitive accuracy compared to reID baselines, while this CNN has less than 40\% parameters of the reID baselines. By pre-training this CNN on ImageNet for initialization, we achieve state-of-the-art performance on three reID benchmarks with only half the number of parameters.
\end{itemize}

\section{Related Work}\label{sec:related-work}

{\bf Person reID.} Prevailing algorithms have achieved great success with the deep learning technique~\cite{xiao2016learning,chen2017beyond,saquib2018pose,Huang_2018_AOS,Sun_2018_PCB,fan18_unsupervisedreid,ding2019adaptive}.
Xiao~et~al.~\cite{xiao2016learning} propose a pipeline to deep feature representations from multiple datasets.
Chen~et~al.~\cite{chen2017beyond} design a quadruplet loss to make deep CNN capture both inter-class and intra-class variations.
Saquib~et~al.~\cite{saquib2018pose} take the body joint maps as additional inputs to enable deep CNN to learn pose sensitive representations.
Sun~et~al.~\cite{Sun_2018_PCB} leverage a part-based CNN model and a refined part pooling method to learn discriminative part-informed features.

On the one hand, these deep-based reID algorithms~\cite{xiao2016learning,Su_2017_PDC,saquib2018pose,Sun_2018_PCB,suh2018part} heavily rely on the classification CNN backbones, such as VGG~\cite{simonyan2015very}, Inception~\cite{Szegedy_2015_googlenet}, and ResNet~\cite{he2016resnet}.
These CNN backbones are specifically designed for the classification problem and experimented on classification datasets, which may not align with reID and limit the performance of reID algorithms.
On the other hand, they incorporate reID specific domain knowledge to boost the classic CNN models, such as part cues~\cite{Sun_2018_PCB,suh2018part}, pose~\cite{saquib2018pose}, and reID specific loss~\cite{chen2017beyond,HermansBL17_TriNet}.
In this work, we not only inherit the merit of previous reID methods but also overcome their disadvantages.
We automatically find a reID specific CNN architecture over a reID search space.

{\bf Neural Architecture Search.} Our work is motivated by recent researches on NAS~\cite{liu2019darts,ZophL2017NASNet,ZophVSL2018LTA,Andrew2017SMASH,Chen_2019_CVPR,dong2019search,fang2019densely,fang2019densely,Chen_2019_CVPR}, while we focus on searching for a reID model with high performance instead of a classification model.
Most of NAS approaches~\cite{liu2019darts,ZophL2017NASNet,ZophVSL2018LTA,Andrew2017SMASH,Hieu2018ENAS} search CNN on a small proxy task and transfer the found CNN structure to another large target task.
Zoph et~al.~\cite{ZophL2017NASNet,ZophVSL2018LTA} apply reinforcement learning to search CNN, while the search cost is more than hundreds of GPU days.
Real~et~al.~\cite{Real2019RECAS} modify the tournament selection evolutionary algorithm by introducing an age property to favor the younger CNN candidates.
Brock~et~al.~\cite{Andrew2017SMASH} and Bender~et~al.~\cite{bender2018understanding} explore the one-shot NAS approaches.
Liu~et~al.~\cite{liu2019darts} relax the discrete search space so as to search CNN in a differentiable way.
Dong~et~al.~\cite{dong2019search} propose a differentiable sampling approach to improve \cite{liu2019darts}.
Benefited from parameter sharing technique~\cite{Hieu2018ENAS,liu2019darts}, we discard the proxy paradigm and directly search a robust CNN on the target reID dataset.
Besides, previous NAS algorithms~\cite{liu2019darts,ZophL2017NASNet,ZophVSL2018LTA,Andrew2017SMASH,Hieu2018ENAS,bender2018understanding} focus on the classification problem. They are generic and can be readily applied to the reID problem.
However, without considering reID specific information, such as semantics~\cite{kalayeh2018human}, occlusion~\cite{Huang_2018_AOS}, pose~\cite{saquib2018pose}, and part~\cite{Sun_2018_PCB}, generic NAS approaches can not guarantee that the searched CNN is suitable for reID tasks.
In this work, based on an efficient NAS algorithm~\cite{liu2019darts}, we adopt two techniques to modify it for the reID problem.
We modified the objective function and training strategy to adapt to the reID problem.
In addition, we design a part-aware module and integrate it into the standard NAS search space, which could allow us to find a better CNN and advance the study of the NAS search space.

\section{Methodology}

In this section, we will show how to search for a reID CNN with high performance.
We will first introduce the preliminary background of NAS in Sec.~\ref{sec:preliminaries}.
Then we propose a new search algorithm for reID, introduced in Sec.~\ref{sec:reID-search}.
Furthermore, we design a new reID search space in Sec.~\ref{sec:part-aware-reID}, which integrates our proposed part-aware module and the standard NAS search space.
Lastly, we discuss some future directions for reID in Sec.~\ref{sec:discussion}.

\subsection{Preliminaries}\label{sec:preliminaries}

Most NAS approaches stack multiple copies of a neural cell to construct a CNN model~\cite{ZophVSL2018LTA,Liu_2018_PNAS,Real2019RECAS}. A neural cell consists of several different kinds of layers, taking output tensors from previous cells and generating a new output tensor.
We follow previous NAS approaches~\cite{ZophL2017NASNet,ZophVSL2018LTA,liu2019darts} to search for the topology structure of neural cells.

Specifically, a neural cell can be viewed as a directed acyclic graph (DAG) with $B$ blocks. Each block has three steps: (1) take two tensors as inputs, (2) apply two operations on these two tensors, respectively, (3) sum these two tensors. The applied operation is selected from an operation candidate set $\gO$. Following some previous works~\cite{Hieu2018ENAS}, we use the following operations in our $\gO$:
(1) 3$\times$3 max pooling, (2) 3$\times$3 average pooling, (3) 3$\times$3 depth-wise separable convolution, (4) 3$\times$3 dilated convolution, (5) zero operation (none), (6) identity mapping.
The $i$-th block in the $c$-th neural cell can be represented as a 4-tuple, i.e., ($I_{i1}^{c}$, $I_{i2}^{c}$, $\gO_{i1}^{c}$, $\gO_{i2}^{c}$). Besides, the output tensor of the $i$-th block in the $c$-th neural cell is:
\begin{align}\label{eq:basic-block}
    I_{i}^{c} = \gO_{i1}^{c}( I_{i1}^{c} ) + \gO_{i2}^{c}( I_{i2}^{c} ),
\end{align}
\noindent where $\gO_{i1}^{c}$ and $\gO_{i2}^{c}$ are selected operations from $\gO$ for the $i$-th block. $I_{i1}^{c}$ and $I_{i2}^{c}$ are selected from the candidate input tensors $\gI_{i}^{c}$, which consists of output tensors from the last two neural cells ($I^{c\textrm{-}1}$ and $I^{c\textrm{-}2}$) and output tensors from the previous block in the current cell.

To search for the choices of $\gO_{i1}^{c}$ and $I_{i1}^{c}$ in Eq.~\eqref{eq:basic-block}, we relax the categorical choice of a particular operation as a softmax over all possible operations following~\cite{liu2019darts}:
\begin{align}\label{eq:relax-op}
    \gO_{i1}^{c}(I_{i1}^{c}) = \sum_{H\in\gI_{i}^{c}} \sum_{o\in\gO} \frac{ \exp(\alpha_{o}^{(H,i)}) }{\sum_{o'\in\gO} \exp(\alpha_{o'}^{(H,i)}) } o(H),
\end{align}
\noindent where $\alpha = \{ \alpha_{o}^{(H,i)} \}$ represents the topology structure for a neural cell, named as \textbf{\textit{architecture parameters}}. Denote the parameters of all operations in $\gO$ as $\omega$, named as \textbf{\textit{operation parameters}}, a typical differentiable NAS approach~\cite{liu2019darts} jointly trains $\omega$ on the training set and $\alpha$ on the validation set. After training, the strength of $H$ to $I_{i}^{c}$ is defined as $\max_{o\in\gO,o\neq\textrm{none}}\frac{\exp(\alpha_{o}^{(H,i)})}{\sum_{o'\in\gO} \exp(\alpha_{o'}^{(H,i)})}$. The $H\in\gI_{i}^{c}$ with the maximum strength is selected as $I_{i1}^{c}$, and the operation with the maximum weight for $I_{i1}^{c}$ is selected as $\gO_{i1}^{c}$.
This paradigm is designed for the classification problem.
Inspired from them, we apply several improvements to adapt this paradigm into the person reID problem.

\begin{algorithm}[!t]
\small
\caption{The Auto-ReID Algorithm}
\label{algo:pseudocode}
\begin{algorithmic}
\Require the architecture parameter $\alpha$ and the operation parameter $\omega$; the training set $\sD_{T}$ and the evaluation set $\sD_{E}$; a class-balance data sampler; \\
\textbf{1:} Split $\sD_{T}$ into the search training set $\sD_{train}$ and the search validation set $\sD_{val}$
\While{not converged} \\
\hspace{1mm}\textbf{2:} Use the sampler to get batch data from $\sD_{train}$ \\
\hspace{1mm}\textbf{3:} Update $\omega$ via the retrieval loss in Eq.~\eqref{eq:mixture} \\
\hspace{1mm}\textbf{4:} Use the sampler to get batch data from $\sD_{val}$ \\
\hspace{1mm}\textbf{5:} Update $\alpha$ via the retrieval loss in Eq.~\eqref{eq:mixture}
\EndWhile \\
Obtain the final CNN from $\alpha$ following the strategy in~\cite{liu2019darts} \\
Optimize this CNN on the training set $\sD_{T}$ by the standard reID training strategy \\
Evaluate the trained CNN on the evaluation set $\sD_{E}$
\end{algorithmic}
\end{algorithm}

\begin{figure*}[!th]
\centering
\includegraphics[width=\linewidth]{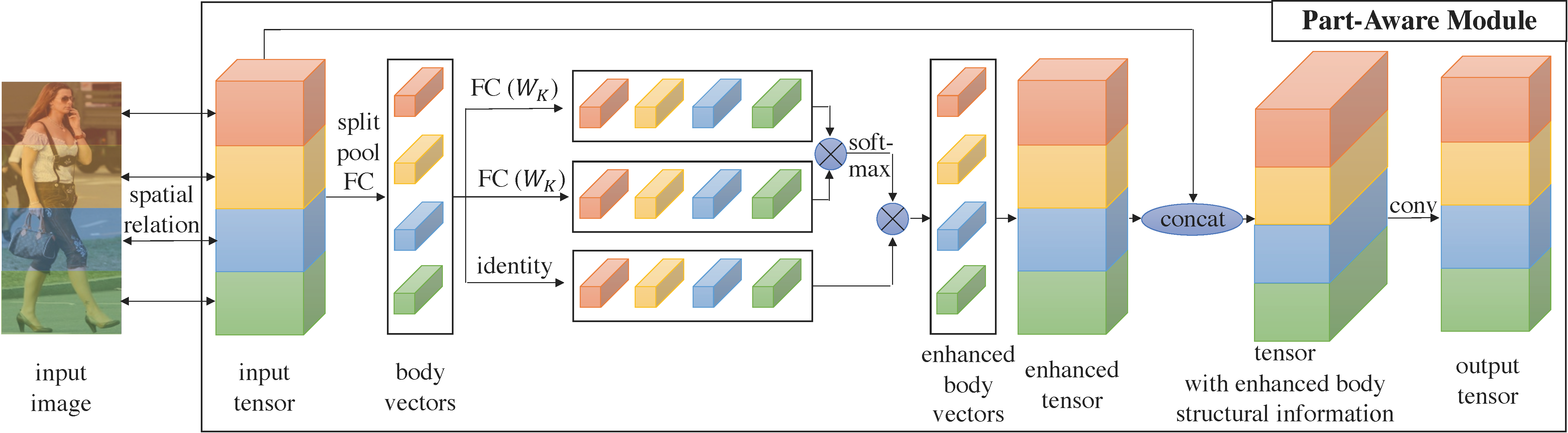}
\caption{
The proposed part-aware module for the reID search space.
Given a pedestrian feature tensor, this module can integrate human body structural cues into the input tensor.
It first vertically splits the input feature tensor into $M=4$ body part features, and then averages each part tensor into a vector and uses a linear layer to transform each of them into a new part feature vector, denoted as ``body vectors''.
These $M$ part vectors are interacted via a self-attention mechanism, and each part vectors could include more body part specific information.
Later, these $M$ vectors are repeated and concatenated to recover them into the same spatial shape as the input tensor, named as ``enhanced tensor''.
Finally, we fuse this global feature tensor and the original input tensor by a one-by-one convolutional layer.
}
\label{fig:part-aware-module}
\end{figure*}

\subsection{ReID Search Algorithm}\label{sec:reID-search}

Prevailing NAS approaches focus on searching for a well-performed architecture in the classification task, in which the softmax with cross-entropy loss is applied to optimizing both $\alpha$ and $\omega$~\cite{liu2019darts,ZophVSL2018LTA}.
In contrast, reID tasks aim to learn a discriminative feature extractor during training, so that the extracted feature can retrieve images of the same identity during evaluation. Simply inheriting the cross-entropy loss can not guarantee a good retrieve performance. We need to incorporate reID specific knowledge into the searching algorithm.

\textbf{Network Structure}.
We use the macro structure of ResNet~\cite{he2016resnet} for our reID backbone, where each residual layer is replaced by a neural cell. We search the topology structure of neural cells.
Denote the feature extracted from the backbone as $\vf$, we use one embedding layer to transfer the feature $\vf$ into $\vg$ following~\cite{Sun_2018_PCB}, and we use another linear transformation layer to map the feature $\vg$ into the logits $\vh$ with the output dimension of $C$, where $C$ denotes the number of training identities. Two dropout layers are added between $\vf\&\vg$ and $\vg\&\vh$, respectively.

\textbf{Objective}.
The classification model usually applies the softmax with cross-entropy loss on $\vh$ as follows:
\begin{align}\label{eq:softmax}
  L_{s} = \sum_{i=1}^{N} - \log \frac{\exp(\vh_{i}[{c}])}{\sum_{c'=1}^{C}\exp(\vh_{i}[{c'}])} ,
\end{align}
\noindent where $\vh_{i}$ indicates the feature $\vh$ of the $i$-th sample, and $\vh_{i}[{c}]$ indicates the $c$-th element in $\vh_{i}$.
$N$ is the number of samples during training.
The reID model usually applies the triplet loss as:
\begin{align}\label{eq:triplet}
  L_{t} = \sum_{i=1}^{N} \max(\textrm{margin}, || \vf_{i} - \vf_{i}^{p} || - || \vf_{i} - \vf_{i}^{n} ||),
\end{align}
\noindent where $\vf_{i}$ indicates the feature $\vf$ of the $i$-th sample. $\vf_{i}^{p}$ indicates the hardest positive feature of $\vf_{i}$.
The $\textrm{margin}$ term indicates the margin of triplet loss.
In another word, $\vf_{i}^{p}$ is another feature with the maximum Euclidean distance of $\vf_{i}$ and the same identity of $\vf_{i}$ in one batch.
$\vf_{i}^{n}$ is the hardest negative feature of $\vf_{i}$. In another word, $\vf_{i}^{n}$ is another feature with the minimum Euclidean distance of $\vf_{i}$ and the different identity of $\vf_{i}$ in one batch.
Since the triplet loss is sensitive to the batch data, we should carefully sample training data in each batch.
We adopt a class-balance data sampler to sample batch data for triplet loss.
This sampler first samples uniformly sample some identities, and then, for each identity, it randomly sample the same number of images. 
To align with the reID problem and leverage the mutual benefit from the cross-entropy and triplet losses, we consider a mixture retrieval loss of $L_{s}$ and $L_{t}$ as follows:
\begin{align}\label{eq:mixture}
    L_{ret} = \lambda L_{s} + (1-\lambda) L_{t},
\end{align}\noindent where $\lambda \in [0,1]$ is a weight balancing of $L_{s}$ and $L_{t}$.

We show our overall algorithm (Auto-ReID) in Alg.~\ref{algo:pseudocode}, which solves a bi-level optimization problem~\cite{dong2019search,liu2019darts,liu2019ppn,ZophL2017NASNet}.
We first search for a robust reID model by alternatively optimizing $\alpha$ with $L_{t}$ and $\omega$ with $L_{ret}$.
The searched CNN is derived from $\alpha$ based on the same strategy as in~\cite{liu2019darts,dong2019search}.
After we find a robust CNN for the reID task, we train and evaluate this CNN in the standard way.

\begin{table*}[!t]
    \setlength{\tabcolsep}{9pt}
	\centering
    \begin{tabular}{|l|c|c|c|c|c|c|}
    \hline
    \multicolumn{1}{|c|}{Architectures} & mAP & Rank-1 & Rank-5 & Rank-10 & Params~(M) & FLOPs~(G) \\
    \hline\hline
    ResNet-18~\cite{he2016resnet} & 66.0 & 85.2 & 94.6 & 96.5 & 11.6 & 1.7 \\
    ResNet-34~\cite{he2016resnet} & 68.0 & 86.7 & 94.8 & 96.6 & 21.7 & 3.4 \\
    ResNet-50~\cite{he2016resnet} & 68.5& 87.2&95.5&97.1&25.1&3.8\\
    \hline\hline
    \multicolumn{1}{|l|}{DARTS~\cite{liu2019darts}}&65.2& 85.6 & 94.3 & 96.4 & 9.1 & 1.7 \\
    GDAS~\cite{dong2019search} & 66.8&86.5&94.7&96.9 &13.5 & 2.3\\\hline
    \multicolumn{1}{|l|}{Baseline (run 1)}   & 68.5          & 87.0 & 95.4 & 97.1 & 11.9  & 2.0  \\
    \multicolumn{1}{|l|}{Baseline (run 2)}   & 68.2          & 86.8 & 95.6 & 97.3 & 10.8  & 1.9\\
    \multicolumn{1}{|l|}{Baseline (run 3)}   & 65.8          & 85.8 & 95.0 & 96.8 & 8.6  & 1.6\\
    \multicolumn{1}{|l|}{Baseline (run 4)}   & 66.5          & 86.5 & 95.0 & 96.9 & 9.0  & 1.7\\\hline
    \multicolumn{1}{|l|}{Baseline + ReID Search Space (run 1)}  & 71.3          & 90.0 & 96.5 & 97.7 & 10.9 & 1.7  \\
    \multicolumn{1}{|l|}{Baseline + ReID Search Space (run 2)}  & 71.2          & 89.1 & 96.1 & 97.5 & 14.8 & 2.2  \\
    \multicolumn{1}{|l|}{Baseline + ReID Search Space (run 3)}  & 74.6          & 90.7 & 96.9 & 98.1 & 13.1 & 2.1  \\
    \multicolumn{1}{|l|}{Baseline + ReID Search Space (run 4)}  & 72.2          & 89.3 & 96.6 & 97.8 & 14.6 & 2.0  \\\hline
    \multicolumn{1}{|l|}{Retrieval + ReID Search Space (run 1)} & 72.7 & 89.7 & 96.7 & 98.0 & 11.4 & 1.8  \\
    \multicolumn{1}{|l|}{Retrieval + ReID Search Space (run 2)} & 73.4 & 90.2 & 96.4 & 97.7 & 14.5 & 2.3  \\
    \multicolumn{1}{|l|}{Retrieval + ReID Search Space (run 3)} & 73.1 & 89.5 & 96.6 & 97.9 & 13.1 & 2.0  \\
    \multicolumn{1}{|l|}{Retrieval + ReID Search Space (run 4)} & 74.2 & 90.3 & 96.6 & 97.9 & 14.1 & 2.1  \\
    
    \hline
\end{tabular}
\vspace{2mm}
\caption{
We analyze the effect of each component in our proposed method. All CNN models are trained in the same strategy and do \textbf{not} use ImageNet pre-training for initialization. We search four times for each search algorithm and show their results.
During searching, we use $C$=32 and $\vl$=[2,2,2,2] to improve the efficiency.
To train the searched architecture, we use $C$=64 and $\vl$=[2,2,2,2] to keep the number parameters similar as ResNet-18.
}
\vspace{-2mm}
\label{table:comparison_resnet}
\end{table*}

\subsection{ReID Search Space with Part-Aware Module}\label{sec:part-aware-reID}

The search space covers all possible candidate CNNs to be found, which is important for NAS.
A standard search space in NAS is ``NASNet search space''~\cite{ZophVSL2018LTA}, which contains different kinds of convolutional layers, different kinds of pooling layers, etc.
None of these layers can explicitly handle pedestrian information, which requires a delicate design and some unique operations. In this paper, we take the first step to explore a search space that fits the reID problem.

Motivated by the fact that body part information can improve the performance of a reID model~\cite{Sun_2018_PCB,Su_2017_PDC}, we design a part-aware module and combine it with a common search space ($\gO$) to construct our reID search space $\gO_{reid}$:
(1) \textbf{\textit{part-aware module}}, (2) $3\times3$ max pooling, (3) $3\times3$ average pooling, (4) $3\times3$ depth-wise separable convolution, (5) $3\times3$ dilated convolution, (6) zero operation, and (7) identity mapping.

The part-aware module is shown in Fig.~\ref{fig:part-aware-module}.
Given an input feature tensor $\mF$, we first split it into $M$ parts vertically, where we show an example of $M=4$ in Fig.~\ref{fig:part-aware-module}.
After we obtain the part features, we average pool each part feature over the spatial dimension and apply a linear transformation to the pooled features, and can thus obtain $M$ local body part feature vectors.
Then, we apply a self-attention mechanism~\cite{vaswani2017attention} on these $M$ part feature vectors.
In this way, we can incorporate global information into each part vectors to enhance its body structure cues.
Later, we repeat each part vector into its original spatial shape and 
and concatenate the repeated part features vertically into a body structure enhanced feature tensor.
Finally, we fuse this part-aware tensor and the original input feature tensor via channel-wise concatenate way, and apply a one-by-one convolutional layer on this fusion tensor to generate the output tensor.
Our designed part-aware module can capture useful body part cues and integrate this structural information into the input features.
Besides, the parameter size and number of calculation of the proposed part-aware module are similar to the 3x3 depth-wise separable convolution, and thus will not affect the efficiency of the found CNN compared with using a standard NAS search space.

\subsection{Discussion}\label{sec:discussion}

Researchers have trend to move their focus from manually architecture design to automated architecture design in many areas, e.g., classification~\cite{ZophVSL2018LTA,liu2019darts,dong2019search} and segmentation~\cite{chen2018searching}.
In the reID community, the breakthrough of the reID performance is usually benefited from improvements on the CNN architecture.
We present the first effort towards applying automated machine learning to reID.
After so many different architectures proposed for reID~\cite{xiao2016learning,Su_2017_PDC,saquib2018pose,Sun_2018_PCB,suh2018part}, it becomes more difficult to manually find a better architecture.
It is time to automatically design a good reID architecture, and to our knowledge, this is the first time that an automated algorithm has matched state-of-the-art performance using architecture search techniques.

\section{Experiments}\label{sec:exps}
We empirically evaluate the proposed method in this section.
We will first introduce the used datasets in Sec.~\ref{sec:exps-dataset} and implementation details in Sec.~\ref{sec:exps-implementation}.
Then, we will ablatively study different aspects of our Auto-ReID algorithm in Sec.~\ref{sec:exps-ablation}, and also compare the CNN found by our approach with other state-of-the-art algorithms in Sec.~\ref{sec:exps-SOTA}.
Lastly, we make some qualitative analysis in Sec.~\ref{sec:exps-vis}.

\subsection{Datasets and Evaluation Metrics}\label{sec:exps-dataset}

\textbf{Market-1501}~\cite{zheng2015scalable} is a large-scale person reID dataset which contains 19,372 gallery images, 3,368 query images and 12,396 training images collected from six cameras. There are 751 identities in training set and 750 identities in the test set and they have no overlap. Every identity in the training set has 17.2 images on average.

\textbf{CUHK03}~\cite{li2014deepreid} consists of 1,467 identities and 28,192 bounding boxes. There are 26,264 images of 1,367 identities are used for training and 1,928 images of 100 identities are used for testing. We use the new protocol to split the training and test data as proposed by~\cite{Zhong_2017_CVPR}.

\textbf{MSMT17}~\cite{wei2018msmt17} is currently the largest person reID dataset, which contains 126,441 images of 4,101 identities in 15 cameras. This dataset is composed of the training set, which contains 32,621 bounding boxes of 1,041 identities and the test set including 93,820 bounding boxes of 3,060 identities. From the test set, 11,659 images are used as query images and the other 82,161 bounding boxes are used as gallery images. This challenging dataset has more complex scenes and backgrounds, e.g., indoor and outdoor scenes, than others.

\textbf{Evaluation Metrics.} To evaluate the performance of our Auto-ReID and compare with other reID methods, we report two common evaluation metrics: the cumulative matching characteristics (CMC) at rank-1, rank-5 and rank-10 and mean average precision (mAP) on the above three benchmarks following the common settings~\cite{zheng2015scalable,wei2018msmt17}.

\subsection{Implementation Details}\label{sec:exps-implementation}

\textbf{Search Configurations.}
During the searching period, we randomly select 50\% images from official training set as the search training set $\sD_{train}$ and other images as the search validation set $\sD_{val}$.
We choose the ResNet macro structure to construct the overall network.
This network has a 3x3 convolutional head and four blocks sequentially, where each block has several neural cells. We denote the number of cells in each block is $\vl_{1}$, $\vl_{2}$, $\vl_{3}$, and $\vl_{4}$. We denote $\vl=[\vl_{1}, \vl_{2}, \vl_{3}, \vl_{4}]$. We denote the channel of the first convolutional layer as $C$, and each block will double the number of channels.
The first cell in the $2$-th, $3$-th, and $4$-th block is a reduction cell, and other cells are the normal cell~\cite{Liu_2018_PNAS,liu2019darts}.
By default, we use $C$=32 and $\vl=[2,2,2,2]$ to search a suitable CNN architecture.

During searching, we use a input size of $384\times128$, a batch size of 16, the total epoch of 200.
We use momentum SGD to optimize $\omega$ with the initial learning rate of 0.1 and decrease it to 0.001 in a cosine scheduler. The momentum for SGD is set as 0.9.
We use Adam to optimize $\alpha$ with the initial learning rate of 0.02, which is decayed by 10 at 60-th and 150-th epoch.
The weight decay for both SGD and Adam is set as 0.0005.
The $\textrm{margin}$ is set as 0.3 when using the retrieval objective.
The $\lambda$ is set as 0.5.

In experiments, ``DARTS'' and ``GDAS'' in Table~\ref{table:comparison_resnet} denotes that we train the network provided in \cite{liu2019darts,dong2019search} on Market-1501. ``Baseline'' indicates we use DARTS (first order~\cite{liu2019darts}) to search an architecture on Market-1501, and then train the searched model. We use ``Baseline + ReID Search Space'' to denote the baseline searching algorithm on the proposed reID search space. And we use ``Retrieval + ReID Search Space'' to denote the proposed retrieval-based searching algorithm on the proposed reID search space.
``Retrieval + ReID Search Space'' costs about 1 day to finish one searching procedure on a single NVIDIA Tesla V100.

\textbf{Training Configurations.}
In the training phrase, we use an input size of 384$\times$128, $C$=64, and $\vl=[2,2,2,2]$.
Following previous works, we use random horizontal flipping and cropping for data augmentation. We set $\lambda$ as 0.5 for the retrieval loss.
For training from scratch, in one batch, our class-balance data sampler will first random select 8 identities and then random sample 4 images for each identity.
When using ImageNet pre-trained models, it randomly samples 16 identities and then samples 4 images for each identity.
We train the model for 240 epochs, using Adam as the optimizer with a momentum of 0.9 and a weight decay of 0.0005. We start the learning rate from 0.0035 and decay it by 10 at the 80-th and 150-th epochs.

\begin{table}[!t]
     \setlength{\tabcolsep}{4.5pt}
	 \centering
     \begin{tabular}{|c|c|c|c|c|c|}
     \hline
     \multicolumn{3}{|c|}{Configurations} & \multirow{3}{*}{Rank-1} &   \multirow{3}{*}{mAP} & \multirow{3}{*}{\tabincell{c}{Params \\ (M)}} \\
     \cline{1-3}\multirow{2}{*}{Search} & \multicolumn{2}{|c|}{Train} & & & \\ \cline{2-3}
     &$C$&$\vl$& & &\\\hline
     \multirow{ 4}{*}{\tabincell{c}{$C$=16 \\ $\vl$=[2,2,2,2]}} & \multirow{2}{*}{16} &[2,2,2,2] & 81.2 & 58.9 & 1.1 \\ 
     & &[3,4,6,3]& 79.5 & 53.5 & 1.7 \\\cline{2-6}
     &\multirow{2}{*}{32}&[2,2,2,2]& 86.4 & 66.0 & 3.8 \\
     & &[3,4,6,3]& 85.4 & 62.7 & 5.9 \\\hline
     \multirow{ 4}{*}{\tabincell{c}{$C$=16 \\ $\vl$=[3,4,6,3]}} & \multirow{2}{*}{16} & [2,2,2,2] & 77.3 & 54.0 & 1.0 \\
     & &[3,4,6,3]& 81.2 & 56.2 & 1.4 \\\cline{2-6}
     &\multirow{2}{*}{32} &[2,2,2,2]& 85.9 & 66.0 & 3.1 \\
     &&[3,4,6,3]& 87.2 & 66.5 & 4.9 \\\hline
     \multirow{ 4}{*}{\tabincell{c}{$C$=32 \\ $\vl$=[2,2,2,2]}} & \multirow{2}{*}{16} & [2,2,2,2] & 80.7 & 58.3 & 1.2 \\
     & &[3,4,6,3]& 80.3 & 56.7 & 1.9 \\\cline{2-6}
     &\multirow{2}{*}{32}&[2,2,2,2]& 87.6 & 68.3 & 4.1 \\
     & &[3,4,6,3]& 85.1 & 64.9 & 6.6 \\\hline
     \multirow{ 4}{*}{\tabincell{c}{$C$=32 \\ $\vl$=[3,4,6,3]}} & \multirow{2}{*}{16}&[2,2,2,2]& 78.0 & 55.4 & 1.2 \\
     & &[3,4,6,3]& 80.0 & 55.5 & 1.8 \\\cline{2-6}
     &\multirow{2}{*}{32}&[2,2,2,2]& 85.6 & 66.2 & 3.9 \\
     & &[3,4,6,3]& 86.5 & 64.6 & 6.2 \\\hline
\end{tabular}
\vspace{2mm}
\caption{
We experiment using different configurations in the searching and the training procedures. Apart from the $C$ and $\vl$, we keep other hyper-parameters the same for different configurations.
All above experiments do not use ImageNet pre-training.
}
\vspace{-2mm}
\label{table:configs}
\end{table}

\subsection{Ablation Study}\label{sec:exps-ablation}

\begin{table}[!t]
\setlength{\tabcolsep}{3.2pt}
\centering
\begin{tabular}{|c|c|c|c c|}
\hline
\multirow{2}{*}{Methods }&{\multirow{2}{*}{Backbone}}&{\multirow{2}{*}{\tabincell{c}{Params \\ (M)}}}&\multicolumn{2}{c|}{Market-1501}\\ 
\cline{4-5}
& & &{R-1}&{mAP}\\
\hline
PAN~\cite{zheng2018PAN} &ResNet50&$>25.1$&82.8&63.3  \\
TriNet~\cite{HermansBL17_TriNet} & ResNet-50 &$25.1$&84.9&69.1       \\
AOS~\cite{Huang_2018_AOS}  & ResNet-50 &$>25.1$&86.4&70.4    \\
MLFN~\cite{Chang_2018_MLFN} &ResNeXt-50&$>25.0$&90.0&74.3    \\
DuATM~\cite{Si_2018_DuATM} &DenseNet-121&$>8.0$&91.4&76.6   \\
PCB~\cite{Sun_2018_PCB} &ResNet-50&$27.2$&93.8&81.6    \\%5.7
Mancs~\cite{Wang_2018_ECCV} &ResNet-50&$>25.1$&93.1&82.3 \\
HPM~\cite{fu2019horizontal} &ResNet-50&25.1&94.2&82.7
\\\hline
Baseline & \multirow{2}{*}{\textit{-}} & 11.9 & 93.8 & 83.4\\ 
\textbf{Auto-ReID} & &13.1&\textbf{94.5}&\textbf{85.1} \\\hline\hline
\multicolumn{5}{|c|}{Using the re-ranking technique~\cite{Zhong_2017_CVPR}.} \\\hline
TriNet~\cite{HermansBL17_TriNet}&ResNet-50& 25.1 & 86.7 &81.1 \\
AOS~\cite{Huang_2018_AOS}&ResNet-50&$>25.1$& 88.7 &83.3 \\
AACN~\cite{Xu_2018_CVPR}&GoogleNet&$>8.0$& 88.7 &83.0\\
PSE+ECN~\cite{saquib2018pose}&ResNet-50&$>25.1$ &90.3& 84.0  \\
PCB~\cite{Sun_2018_PCB} &ResNet-50& 27.2 & 95.1 &91.9 \\\hline
Baseline & \multirow{2}{*}{\textit{-}} & 11.9 & 94.8 & 93.5 \\
\textbf{Auto-ReID} &  & 13.1 & \textbf{95.4} & \textbf{94.2} \\
\hline
\end{tabular}
\vspace{2mm}
\caption{
Comparisons with state-of-art reID models on Market-1501. ``R-1'' indicates the rank-1 accuracy.
}
\vspace{-2mm}
\label{table:market-duke}
\end{table}

To investigate the effect of each component in our Auto-ReID, we perform extensive ablation studies on the Market-1501 dataset. We show the results in Table~\ref{table:comparison_resnet} and Table~\ref{table:configs}.

We compare four searching options in Table~\ref{table:comparison_resnet} without using ImageNet pre-training.
We make several observations:
(1) DARTS~\cite{liu2019darts} and GDAS~\cite{dong2019search} are searched on CIFAR-10, in which the found CNN is worse than a simple reID model based on ResNet-18.
(2) By directly searching on the reID dataset (``Baseline''), we can find better CNNs, which on average have higher mAP and rank-1 accuracy with similar numbers of parameters and FLOPs.
(3) By searching on the proposed reID search space, the performance of the searched CNN can be significantly improved.
(4) Replacing the classification searching loss with the retrieval searching loss, we can obtain slightly better CNNs in most cases.
Though ``Retrieval + ReID Search Space'' finds better CNNs on average compared to ``Baseline + ReID Search Space'', ``Baseline + ReID Search Space (run 3)'' is the architecture with the highest accuracy among all 12 runs.
\textit{Therefore, we use this model as the ``Auto-ReID'' searched model for our following experiments by default.}

\textbf{The effect of different configurations.}
We try different configurations in the searching and training procedure.
We use the ``Baseline + ReID Search Space'' and keep other hyper-parameters the same.
Results are shown in Table~\ref{table:configs}.
First, a higher number of channels during training will yield better accuracy and mAP.
Second, more layers (a larger $\vl$) can result in a better performance only when the value of $\vl$ during searching is the same as the value during training. In another word, a neural cell searched by $\vl=[2,2,2,2]$ is more suitable for an architecture with $\vl=[2,2,2,2]$.
Third, if we use $C$=64 or $\vl$=[3,4,6,3] for experiments in Table~\ref{table:comparison_resnet}, we might find a better CNN. Consider the efficiency, we use a small $C$ and $\vl$ during searching.

\textbf{The effect of formulating the part-aware module as a trainable and operational CNN component.}
We equip NAS-searched models with PCB and show their results in Table~\ref{table:comparisons_d+p}.
Compared to leveraging the part module at the tail, our method searches the most appropriate number and location the for part-aware module. Therefore, we can find a better architecture that is particularly suitable for the reID problem.
As shown in Table~\ref{table:comparisons_d+p}, our method outperforms both ``DARTS+PCB'', ``GDAS+PCB'' and ``Baseline+PCB'' by a large margin of more than 2\% in mAP.

\begin{table}[t!]
\setlength{\tabcolsep}{5pt}
	\centering
    \begin{tabular}{|l|c|c|c|c|}
    \hline
    \multicolumn{1}{|c|}{Architectures} & mAP & Rank-1 & \small{Params}\scriptsize{(M)} & \small{FLOPs}\scriptsize{(G)} \\
    \hline\hline
    DARTS~\cite{liu2019darts} & 65.2& 85.6 & 9.1  & 1.75     \\
    GDAS~\cite{dong2019search}& 66.8& 86.5 & 13.5 & 2.3      \\
    Baseline                  & 68.5 & 87.0 & 11.9 & 2.03 \\\hline
    DARTS+PCB      & 70.3 & 86.9 & 9.1 & 1.75 \\
    GDAS+PCB        & 68.1  & 86.6 & 13.5 & 2.3     \\
    Baseline+PCB     & 70.2 & 86.6 & 11.9 & 2.03\\\hline
    Auto-ReID        & 74.6 & 90.7 & 13.1 & 2.05 \\
    \hline
\end{tabular}
\vspace{2mm}
\caption{
Comparisons with the hand-crafted combination of different CNN backbones and part-related module.
All these models are not pre-trained on ImageNet.
}
\vspace{-2mm}
\label{table:comparisons_d+p}
\end{table}

\subsection{Comparison with State-of-the-art ReID Models}\label{sec:exps-SOTA}

Since all state-of-the-art reID algorithms pre-train their models on ImageNet, we also pre-train our searched CNN on ImageNet for a fair comparison in this section. We train ``Auto-ReID'' with ImageNet initialization on Market-1501 then evaluate the trained model on three reID benchmarks.

\textbf{Results on Market-1501}.
Table~\ref{table:market-duke} compares our method with other state-of-the-art reID models.
Our baseline searching algorithm finds a CNN, which achieves a rank-1 accuracy of 93.8\% and a mAP of 83.4\%, it outperforms other state-of-the-art reID algorithms.
Our Auto-ReID further boosts the performance of ``Baseline''.
The CNN found by our Auto-ReID achieves a rank-1 accuracy of 94.5\% and a mAP of 85.1\%. Note that this CNN reduces the parameters of ResNet-50 based reID models by more than 45\%, whereas it obtains a much higher accuracy and mAP than them.
This experiment demonstrates that our automated architecture search approach can find an efficient and effective model, which successfully removes the noises, redundancies, and missing components in other typical backbones.

Note that our Auto-ReID is orthogonal to other reID techniques, such as re-ranking (RK)~\cite{Zhong_2017_CVPR}, as shown in Table~\ref{table:market-duke}. Using the same augmentation technique~\cite{Zhong_2017_CVPR}, our Auto-ReID also outperforms other reID models.
For example, PCB+RK achieves the mAP of 91.9\%, whereas Auto-ReID+RK achieves the mAP of 94.2\%, which is higher than it by 2.3\%.
Although other techniques can further improve the performance of Auto-ReID, we do not discuss more since it is not the focus of this paper.

\textbf{Results on CUHK03 in Table~\ref{table:comparison_cuhk03}.} There are two types of person bounding boxes: manually labeled and automatically detected. On both settings, our Auto-ReID obtains significantly higher accuracy and mAP than other models.

\textbf{Results on MSMT17 in Table~\ref{table:comparison_msmt17}:}
The previous state-of-the-art method on MSMT17 is PCB,
and our Auto-ReID significantly outperforms it by the mAP of 12\% and the rank-1 accuracy of 10\%.

We made the following observations on three benchmarks:
(1) the CNN model automatically designed by our Auto-ReID outperforms most state-of-the-art reID models on all three datasets.
(2) our Auto-ReID significantly outperforms the baseline searching algorithm.
Although ``Auto-ReID'' is searched on Market-1501, it achieves high accuracy on other two benchmarks.
This suggests that our searched model has a strong transferable ability, which allows us to search over a small proxy dataset when the computational resources are not sufficient to directly search on the target dataset.
Since MSMT17 is much larger than Market-1501, it could be possible to find a better CNN architecture by searching on MSMT17.
When using the default hyper-parameters to search on MSMT17, it yields a slightly worse model compared to the ``Auto-ReID'' model.
It is caused by that different datasets might require different hyper-parameters. With carefully tuned parameters, we believe better CNN could be found using MSMT17.

\textbf{Future Work.} Our Auto-ReID takes the first step to automate the reID model design. The proposed reID search space only considers one possible reID specific module.
More carefully designed basic reID modules can benefit to find a better reID architecture.
We suggest that a reID specific module for NAS is supposed to meet the following requirements: (1) enhancing the human body structure information; (2) eliminating the reID irrelevant (e.g., background) information; (3) being able to take tensors with any shape as input and output tensors with a flexible shape. We would explore this in our future work. In addition, the proposed searching algorithm is a simple extension to the existing NAS algorithm~\cite{liu2019darts}.
We would consider more reID specific knowledge to design more efficient and effective searching algorithms in our future work.

\begin{table}[!t]
	\setlength{\tabcolsep}{7pt}
	\centering
	\begin{tabular}{|c|cc|cc|}\hline
	\multirow{2}{*}{Methods} & \multicolumn{2}{|c|}{Labeled} & \multicolumn{2}{|c|}{Detected}\\\cline{2-5}
		& Rank-1 & mAP & Rank-1 & mAP \\
		\hline

		PAN~\cite{zheng2018PAN}  & 36.9 & 35.0 & 36.3 & 34.0 \\
		SVDNet~\cite{Sun_2017_SVDNet} & 40.9 & 37.8 & 41.5 & 37.3 \\
		HA-CNN~\cite{Li_2018_HA_CNN} & 44.4 & 41.0 & 41.7 & 38.6 \\
		AOS~\cite{Huang_2018_AOS} & -    &   -  & 47.7 & 43.3 \\		
		MLFN~\cite{Chang_2018_MLFN} & 54.7 & 49.2 & 52.8 & 47.8 \\
		PCB~\cite{Sun_2018_PCB} & - & - & 63.7 & 57.5 \\
		Mancs~\cite{Wang_2018_ECCV} & 69.0 & 63.9 & 65.5 & 60.5 \\
		DG-Net\cite{Zhengzhedong_2019_CVPR} & - & - & 65.6 & 61.1 \\
		\hline
		Baseline & 75.0 & 70.1 & 70.5 & 66.5 \\
		\textbf{Auto-ReID} & \textbf{77.9} & \textbf{73.0} & \textbf{73.3} & \textbf{69.3} \\ 
		\hline
	\end{tabular}
	\vspace{2mm}
	\caption{
	Comparison of accuracy and mAP with the state-of-the-art reID models on CUHK03. Note that we use the new evaluation protocol reported in~\cite{Zhong_2017_CVPR}.
	}
	\vspace{-2mm}
	\label{table:comparison_cuhk03}
\end{table}

\begin{table}[!t]
	\centering
	\begin{tabular}{|c|cccc|}
		\hline
		Methods & Rank-1 & Rank-5 & Rank-10 & mAP \\
	    \hline
		GoogleNet~\cite{Szegedy_2015_googlenet}   & 47.6 & 65.0 & 71.8 & 23.0 \\
		PDC~\cite{Su_2017_PDC}  & 58.0 & 73.6 & 79.4 & 29.7 \\
		GLAD~\cite{Wei:2017:GLAD} & 61.4 & 76.8 & 81.6 & 34.0 \\ 
		PCB~\cite{Sun_2018_PCB} & 68.2 & 81.2 & 85.5 & 40.4 \\
		\hline
		Baseline & 74.7 & 86.1 & 89.5 & 48.2 \\
		\textbf{Auto-ReID} & \textbf{78.2} & \textbf{88.2} & \textbf{91.1} & \textbf{52.5} \\
		\hline
	\end{tabular}
	\vspace{2mm}
	\caption{Comparison of accuracy and mAP with the state-of-the-art reID models on MSMT17.
	}
	\vspace{-2mm}
	\label{table:comparison_msmt17}
\end{table}

\subsection{Visualization}\label{sec:exps-vis}

\begin{figure}[!t]
    \centering
    \includegraphics[width=\linewidth]{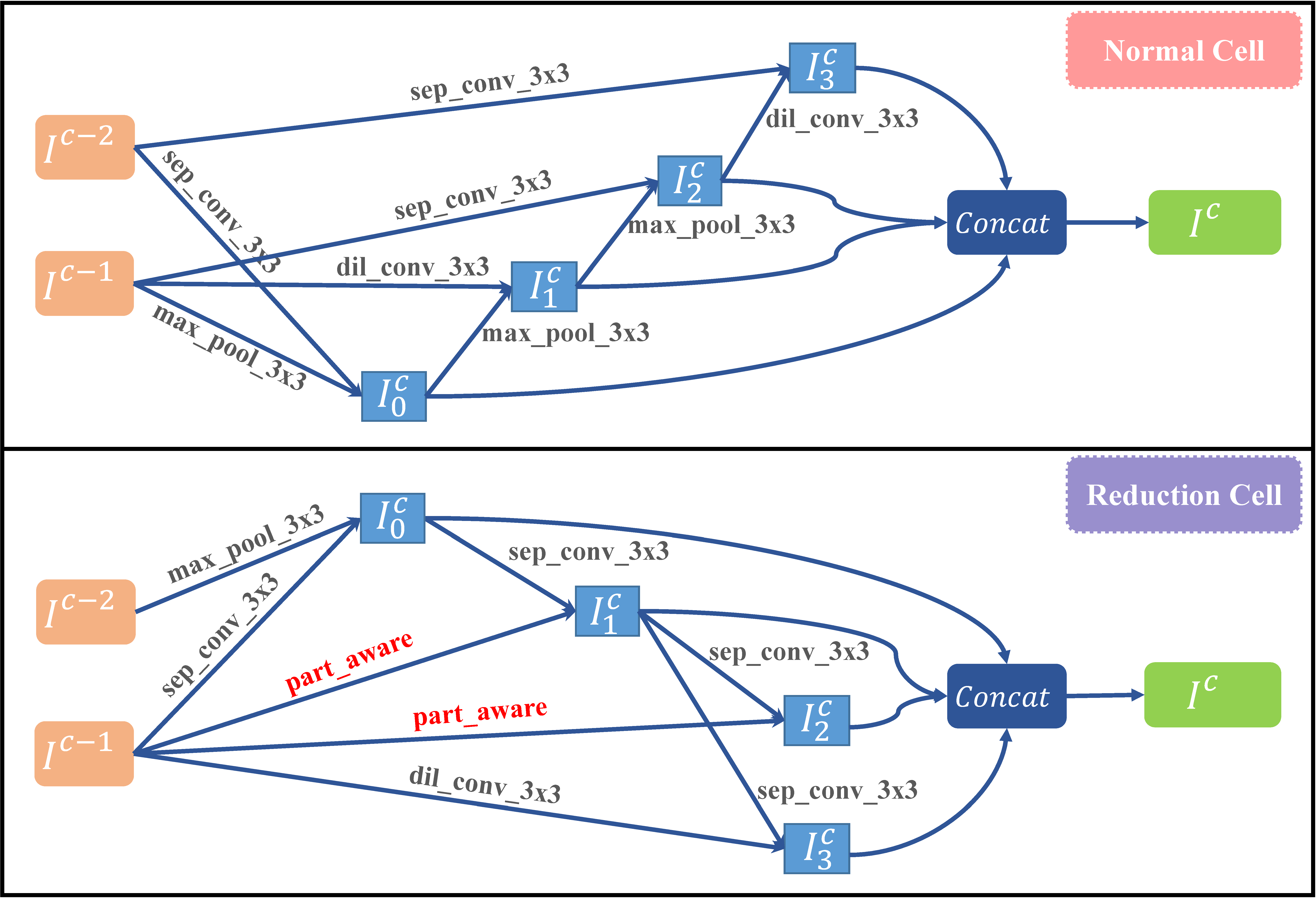}
    \caption{
    The normal cell and the reduction cell used in Table~\ref{table:market-duke}, Table~\ref{table:comparison_cuhk03}, and Table~\ref{table:comparison_msmt17}. This topology structure is complex and hard to designed by human expert.
    }
    \label{fig:vis}
\end{figure}

To better understand what we found during searching, we display one of our searched architectures in Fig.~\ref{fig:vis}.
We show both the normal cell and the reduction cell. These automatically discovered cells are quite complex and are difficult to be found by a human expert with manual tuning. As shown in the reduction cell, two part-aware modules are incorporated in the architecture.
Manually designing a similar architecture as in Fig.~\ref{fig:vis} will cost months, which is inefficient and labor intensive.
This further shows that it is necessary to automate the reID architecture design.

\section{Conclusion}\label{sec:conlusion}
In this paper, we propose an automated neural architecture search for the reID tasks, and we name our method as Auto-ReID.
The proposed Auto-ReID involves a new reID search space and a new retrieval-based searching algorithm.
The proposed reID search space incorporates body structure information into the candidate CNN in the search space. Specifically, it combines a typical classification search space and a novel part-aware module.
Since reID is essentially a retrieval task but current NAS algorithms are merely designed for classification. We equip the NAS algorithm with a retrieval loss, making it particularly suitable for reID.
In experiments, the CNN architecture found by our Auto-ReID significantly outperforms all state-of-the-art reID models on three benchmarks.

{\small
\bibliographystyle{ieee_fullname}
\bibliography{main}
}

\end{document}